\documentclass[journal]{IEEEtran}
\usepackage{amsmath,amsfonts}
\usepackage{algorithmic}
\usepackage{algorithm}
\usepackage{array}
\usepackage[caption=false,font=normalsize,labelfont=sf,textfont=sf]{subfig}
\usepackage{textcomp}
\usepackage{stfloats}
\usepackage{url}
\usepackage{verbatim}
\usepackage{graphicx}
\usepackage{cite}
\usepackage{bm}
\usepackage{booktabs}
\usepackage{multirow}
\usepackage{tabularx}
\usepackage{siunitx}
\usepackage[colorlinks,citecolor=black]{hyperref}
\usepackage{xcolor}
\usepackage{orcidlink}

\begin{document}

\title{Image Manipulation Localization via Adversarial Evidence and Reinforcement Learning Judgment}

\author{Songlin Li\textsuperscript{\large\orcidlink{0009-0008-6793-5624}}, Zhiqing Guo\textsuperscript{\large\orcidlink{0000-0001-6412-334X}},~\IEEEmembership{Member,~IEEE,} Kebin Chen\textsuperscript{\large\orcidlink{0000-0002-4058-5061}}, Wenzhong Yang\textsuperscript{\large\orcidlink{0000-0002-7017-129X}}, Changtao Miao\textsuperscript{\large\orcidlink{0000-0002-7634-9992}}, \\ Gaobo Yang\textsuperscript{\large\orcidlink{0000-0003-2734-659X}}, and Xin Liao, ~\IEEEmembership{Senior Member,~IEEE}
\thanks{This work was supported in part by the Central Government Guides Local Science and Technology Development Fund Projects under Grant ZYYD2026ZY21, in part by the National Natural Science Foundation of China under Grant 62302427, and Grant 62462060. (Corresponding author: Zhiqing Guo)}

\thanks{Songlin Li,  Zhiqing Guo, and Wenzhong Yang are with the School of Computer Science and Technology, Xinjiang University, Urumqi 830046, China. (e-mail: guozhiqing@xju.edu.cn).

Kebin Chen is with Information Support Force Engineering University, Wuhan 430015, China.

Changtao Miao is with the Ant Group, Hangzhou 310000, China.

 Gaobo Yang and Xin Liao are with the College of Computer Science and Electronic Engineering, Hunan University, Changsha 410082, China.
}
}

\markboth{Li \MakeLowercase{\textit{et al.}}: The Courtroom Trial of Pixels}%
{Li \MakeLowercase{\textit{et al.}}: The Courtroom Trial of Pixels}

\IEEEpubid{}

\maketitle
\begin{abstract}
Although some existing image manipulation localization (IML) methods incorporate authenticity-related supervision, this information is typically utilized merely as an auxiliary training signal to enhance the model's sensitivity to manipulation artifacts, rather than being explicitly modeled as localization evidence opposing the manipulated regions. Consequently, when manipulation traces are subtle or degraded by post-processing and noise, these methods struggle to explicitly compare manipulated and authentic evidence, resulting in unreliable predictions in ambiguous areas. To address these issues, we propose a courtroom-style adjudication framework (CSAF) that regards IML task as the confrontation of evidence followed by judgment. The framework comprises a prosecution stream, a defense stream, and a judge model. We first build a dual-hypothesis segmentation architecture on a shared multi-scale encoder, in which the prosecution stream asserts manipulation and the defense stream asserts authenticity. Guided by edge priors, it produces evidence for manipulated and authentic regions through cascaded multi-level fusion, bidirectional disagreement suppression, and dynamic debate refinement. We further develop a reinforcement learning judge model that performs strategic re-inference and refinement on uncertain regions, yielding a manipulated-region mask. The judge model is trained with advantage-based rewards and a soft-IoU objective, and reliability is calibrated via entropy and cross-hypothesis consistency. Experimental results show that our model achieves superior average performance compared with SOTA IML methods. Our code will be released upon acceptance.
\end{abstract}


\section{Introduction}
\IEEEPARstart{I}{mage} manipulation localization (IML) aims to segment manipulated regions at the pixel level. With the rapid advancement of deep learning, numerous IML methods have achieved substantial progress. Existing efforts have primarily advanced IML along two directions: (i) enhancing manipulation-trace modeling through specialized network architectures or auxiliary forensic cues to better capture subtle manipulation-related representations~\cite{wu2019mantra,kwon2021cat,wang2022objectformer,liu2022pscc,li2024image,LiSCAF,guo2025passive}; and (ii) by introducing targeted training objectives to regulate the feature space structure, the discriminability between authentic and manipulated regions is enhanced, thereby enabling the model to better capture subtle and fine-grained manipulation patterns~\cite{zhou2023pre,liu2024attentive,bai2024image,li2025towards,li2026prototype}.

\begin{figure}[!t]
  \centering
  \includegraphics[width=\linewidth]{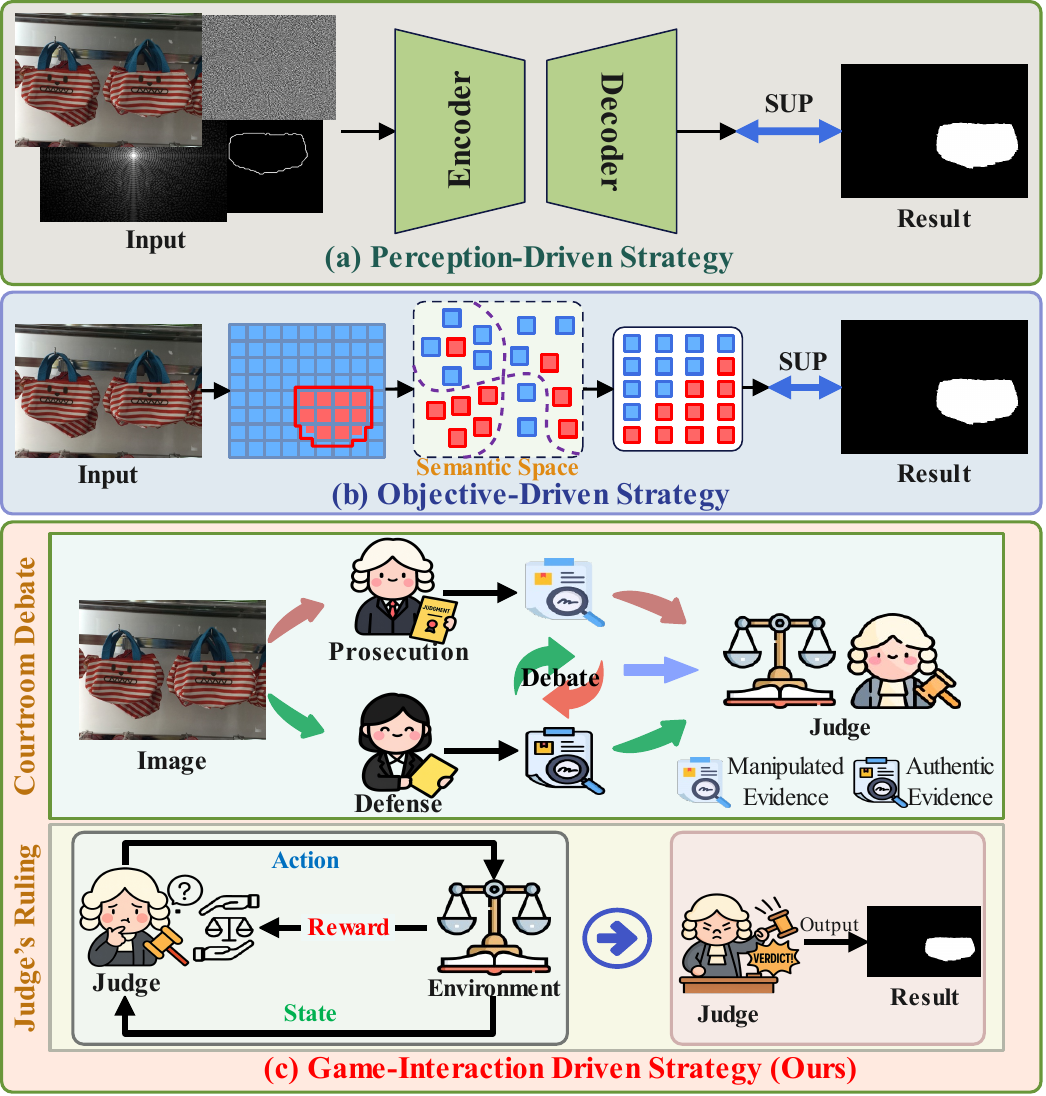}
  \caption{Comparison between our courtroom-style adjudication framework and existing methods. The proposed model consists of two stages: courtroom debate and judge’s ruling.}
  \label{fig_1}
\end{figure}

However, as illustrated in Fig.~\ref{fig_1}(a), methods in the first category fundamentally remain within a manipulation-centric, trace-driven paradigm. They primarily rely on a single stream of evidence supporting the manipulation hypothesis, while rarely exploiting complementary evidence that may support image authenticity. Such one-sided evidence-driven reasoning can become fragile in complex scenarios. Furthermore, as shown in Fig.~\ref{fig_1}(b), although methods in the second category incorporate authenticity-related representations through contrastive learning or feature-level constraints to improve the separability between authentic and manipulated regions, such information is typically used only as an auxiliary training signal rather than being retained at inference time as an independent source of evidence complementary to manipulation cues. Consequently, existing methods still lack an explicit mechanism for comparing and jointly verifying manipulation and authenticity evidence during inference. When manipulation traces are weakened by benign post-processing operations, such as JPEG recompression, image resizing, or social-media transcoding, decisions that rely predominantly on manipulation cues can become unstable. Moreover, the predictive probabilities produced by neural networks do not necessarily reflect their true reliability, and models may therefore remain highly confident even when evidence is insufficient or distributional perturbations lead to incorrect predictions. As a result, a single-pass prediction may be inadequate for reliable localization in regions where forensic evidence is weak, conflicting, or insufficient.

The above limitations can be further understood from the perspectives of competing-hypothesis comparison and uncertainty-aware decision making. A fundamental principle underlying classical hypothesis testing is that reliable decisions should consider the relative support provided by observations for competing hypotheses rather than relying solely on the absolute response associated with a single hypothesis~\cite{neyman1992problem}. This principle motivates us to formulate local IML decisions as a competitive evaluation between a manipulation hypothesis  and an authenticity hypothesis. Meanwhile, prior studies have shown that modern neural networks often suffer from miscalibrated predictive confidence, and that such reliability issues can become more severe under distribution shift~\cite{guo2017calibration,ovadia2019can}. This suggests that an IML model should not only produce a localization prediction, but also identify regions in which the available evidence is insufficient or the resulting decision is unreliable.

Motivated by these principles, we formulate robust IML as a process that combines competing-evidence evaluation with uncertainty-aware re-reasoning. Specifically, we explicitly construct two complementary evidence streams: a prosecution stream that searches for evidence supporting the manipulation hypothesis, and a defense stream that searches for evidence supporting the authenticity hypothesis. When the conclusions derived from the two streams remain consistent after complementary alignment, the corresponding region is supported by mutually consistent evidence. In contrast, substantial disagreement between the two streams indicates potential evidence conflict or predictive uncertainty. Building on this observation, we further introduce a judge model that jointly evaluates cross-hypothesis disagreement and predictive uncertainty, and selectively performs additional reasoning over disputed regions. The courtroom paradigm therefore provides a natural structural interpretation of this process: the prosecution and defense instantiate two competing hypotheses, while the judge performs reliability-aware arbitration and, when necessary, further scrutiny of uncertain evidence. Accordingly, the proposed framework is not merely motivated by a courtroom metaphor, but is grounded in the principles of competing-hypothesis comparison, uncertainty-aware decision making, and adaptive re-reasoning.


To this end, we propose a courtroom-style adjudication framework (CSAF). Built upon a shared multiscale encoder, CSAF adopts a dual-hypothesis segmentation architecture consisting of prosecution and defense streams, which produce evidence for manipulated and authentic regions, respectively. During the evidence formation stage, we develop a dynamic debate mechanism that adaptively suppresses unnecessary cross-stream discrepancies in consensus regions while enhancing informative opposition in disputed regions, thereby refining the evidential representations of both streams. During the judge's ruling stage, a judge model takes the original image together with the prosecution and defense evidence as input. Combined with backbone features, it constructs a dispute representation and local statistics to characterize regions requiring further scrutiny. A policy network then selects refinement actions that guide a lightweight U-Net to re-reason over disputed regions and produce the final manipulation mask. The judge is optimized using advantage-based reinforcement learning (RL) with a soft-IoU reward. Meanwhile, pixel-wise reliability is estimated from predictive entropy and cross-hypothesis consistency. These reliability estimates, together with edge cues, further serve as gating signals for a symmetric Kullback--Leibler (KL) regularizer imposed on the complementary prosecution and defense predictions, which helps mitigate bias and stabilize the final decision.

In summary, our main contributions are as follows:

\begin{itemize}

\item We propose a novel courtroom-style paradigm that formulates IML as a process of adversarial evidence confrontation followed by reliability-aware adjudication. By explicitly modeling both manipulation and authenticity evidence, the proposed framework moves beyond conventional manipulation-centric reasoning and provides a more robust basis for image manipulation localization.

\item We develop a dynamic debate mechanism that adaptively suppresses unnecessary cross-stream discrepancies in consensus regions while enhancing informative opposition in disputed regions, thereby progressively refining the evidence representations of the prosecution and defense streams.

\item We propose an RL-based judge model that performs targeted re-reasoning and error correction in uncertain or conflicting regions. The judge is optimized with an advantage-based soft-IoU reward and further incorporates a reliability-gated symmetric KL regularizer, jointly improving localization accuracy and prediction reliability.

\end{itemize}

\begin{figure*}[t]
  \centering
   \includegraphics[width=\linewidth]{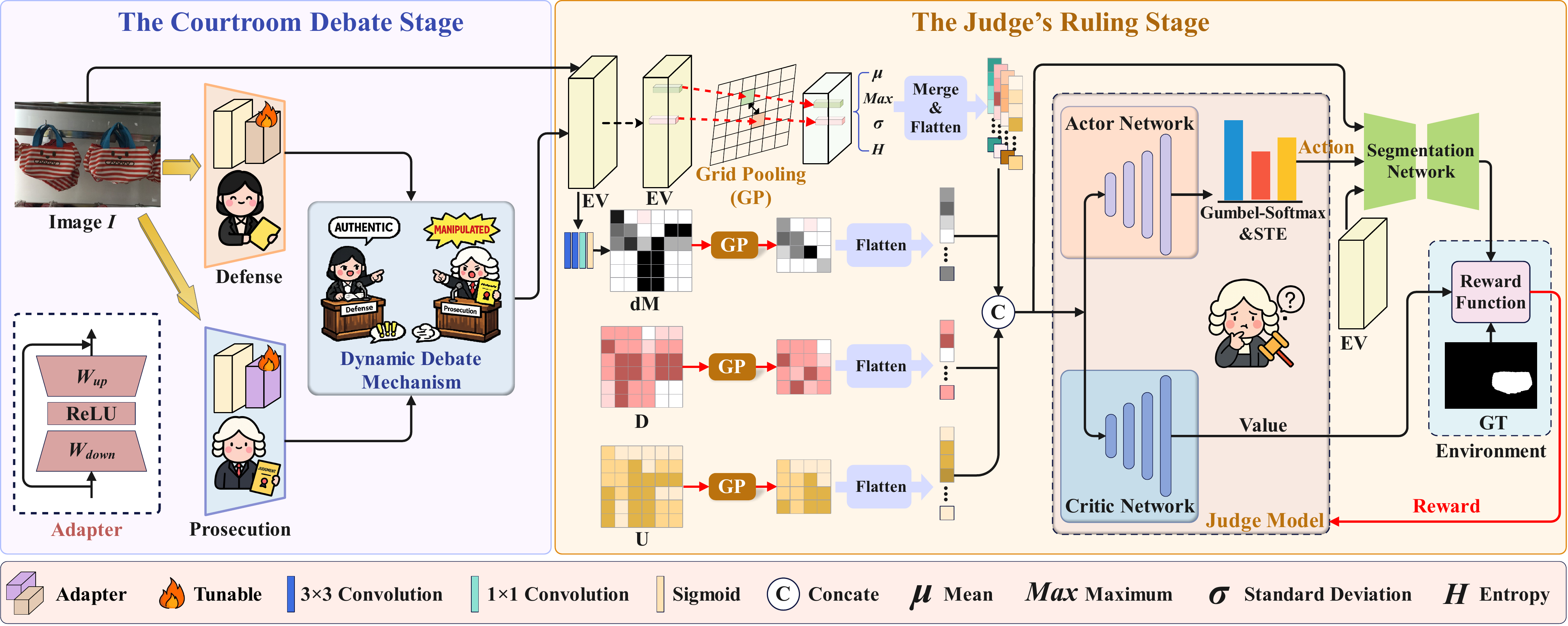}

   \caption{Overview of our courtroom-style adjudication framework, which consists of two main stages: the courtroom debate stage and the judge’s ruling  stage.}
   \label{fig:2}
\end{figure*}

\section{Related Works}
\subsection{Image Manipulation Localization}

\textbf{Perception-Driven Strategy:} 
With the rapid advancement of deep learning, recent Image Manipulation Localization (IML) methods have increasingly adopted specifically designed network architectures, together with complementary visual cues such as edge information, frequency-domain representations, and noise fingerprints, to improve the capability of modeling subtle manipulation traces. Zhuang {\it et al.}~\cite{zhuang2021image} designed a densely connected fully convolutional encoder--decoder network that incorporates dense connections and dilated convolutions to learn discriminative manipulation traces while preserving spatial localization accuracy. Zhuo {\it et al.}~\cite{zhuo2022self} introduced a channel-wise high-pass filtering module and forgery attention into a coarse-to-fine localization framework, together with self-adversarial training, to enhance the robustness of manipulation-sensitive representations. Liu {\it et al.}~\cite{liu2022pscc} proposed PSCC-Net, which progressively models spatial and channel correlations across multiple scales, allowing local manipulation artifacts to be associated with broader contextual information. Dong {\it et al.}~\cite{dong2022mvss} proposed MVSS-Net, which jointly exploits manipulation boundaries and noise-domain artifacts through multi-view learning and multi-scale supervision. Similarly, Guillaro {\it et al.}~\cite{guillaro2023trufor} developed TruFor, which integrates RGB information with a learned noise-sensitive fingerprint, enabling the network to jointly leverage high-level visual semantics and low-level forensic anomalies. Chen {\it et al.}~\cite{chen2024ean} proposed EAN, which enhances subtle IML through boundary-aware modeling and edge feature integration. He {\it et al.}~\cite{ij2024} enhanced IML performance by strengthening the extraction of noise and edge information and incorporating similarity feature constraints.

Despite the diversity of architectures and forensic cues, these methods generally follow a manipulation-centric inference paradigm. Specifically, different representations are mainly extracted and aggregated to strengthen evidence supporting the manipulation hypothesis, while authentic regions are typically treated as implicit counterparts of manipulated regions rather than as an independently modeled source of evidence. Consequently, when manipulation traces become extremely subtle, degraded by post-processing operations, or confused with naturally occurring image irregularities, relying predominantly on manipulation-oriented evidence may result in ambiguous or unreliable predictions.

\textbf{Objective-Driven Strategy:} 
A second research direction follows an objective-driven strategy, where carefully designed learning objectives are introduced to reshape feature distributions and enhance the separability between authentic and manipulated regions. Instead of relying exclusively on predefined forensic traces, these methods aim to learn more discriminative representations by imposing explicit constraints on the feature space. Rao {\it et al.}~\cite{Rao2021} proposed a self-supervised domain adaptation framework for JPEG-robust image forgery localization, which reduces compression-induced domain shifts by simulating JPEG artifacts with a compression approximation network, thereby improving IML accuracy under compressed conditions. Niloy {\it et al.}~\cite{niloy2023cfl} proposed CFL-Net, which introduces contrastive learning into image forgery localization based on the observation that authentic and manipulated regions exhibit different feature distributions. By reducing intra-class variation and enlarging the representation discrepancy between authentic and manipulated regions, CFL-Net improves the generalization capability across different forgery types. However, conventional contrastive formulations face challenges around manipulation boundaries, where individual patches may contain mixed authentic and manipulated pixels. To address this issue, Zhou {\it et al.}~\cite{zhou2023pre} proposed non-mutually exclusive contrastive learning, which dynamically handles boundary patches during representation learning to obtain more discriminative manipulation-sensitive features without relying on additional pre-training data. Lou {\it et al.}~\cite{lou2025exploring} proposed a multi-view pixel-wise learning framework for IML, which captures pixel-level relationships from within-image, cross-scale, and cross-modality perspectives to enhance feature discrimination, generalization, and robustness against post-processing.

These objective-driven approaches represent an important departure from purely trace-driven localization because they explicitly exploit information from both authentic and manipulated regions during representation learning. Nevertheless, authenticity-related information is mainly introduced to optimize feature discrimination during training, rather than being preserved as an independent evidence source for hypothesis verification during inference. As a result, existing methods still lack an explicit mechanism to compare manipulation evidence with authenticity evidence, quantify their disagreement, and perform adaptive re-reasoning when the available evidence is insufficient or conflicting.

Different from the above approaches, our courtroom-style adjudication framework (CSAF) explicitly formulates IML as a competition between two complementary hypotheses. The prosecution stream independently accumulates manipulation evidence, whereas the defense stream searches for evidence supporting authenticity. Instead of using their discrepancy merely as a training objective, CSAF maintains both evidence streams throughout inference and enables their explicit interaction through a debate mechanism. A judge module subsequently performs evidence arbitration over uncertain regions and conducts targeted re-reasoning before producing the final decision. In this way, CSAF extends existing IML beyond manipulation-oriented evidence extraction or authentic--manipulated feature discrimination toward explicit dual-hypothesis evidence arbitration and uncertainty-aware adjudication.

\subsection{Reinforcement Learning}

Reinforcement Learning (RL) provides a sequential decision-making framework, where an agent learns an optimal policy by interacting with the environment and maximizing cumulative rewards under a Markov decision process (MDP). Unlike conventional feed-forward prediction paradigms, RL enables iterative decision-making through state evaluation, action selection, and feedback-driven optimization, offering new opportunities for addressing visual forensic tasks involving subtle, ambiguous, or degraded manipulation traces. However, the exploration of RL in image manipulation localization (IML) remains relatively limited. Existing studies mainly focus on leveraging sequential decision-making mechanisms to optimize localization procedures, while the potential of RL for complex forensic reasoning remains largely unexplored.

Peng et al.~\cite{peng2024employing} formulated pixel-level IML as an MDP, where pixel-wise agents iteratively update forgery probabilities through continuous actions, transforming one-shot prediction into progressive optimization. Cheng et al.~\cite{cheng2025image} further introduced a curiosity-driven deep RL strategy with sparse-reward compensation to enhance exploration and policy learning for IML. These studies demonstrate the potential of RL for iterative localization refinement. Nevertheless, existing RL-based IML methods mainly focus on prediction refinement under a single manipulation hypothesis, where the authenticity hypothesis is not explicitly modeled as a competing source of evidence. Consequently, they lack mechanisms for reasoning over the interaction and competition between manipulation evidence and authenticity evidence. Moreover, when manipulation traces are weakened by image degradation, post-processing operations, or complex content interference, these methods may struggle to assess evidence reliability and conduct targeted corrective reasoning.

Different from these approaches, the proposed model employs RL as an evidence-arbitration mechanism between two complementary hypotheses. The prosecution and defense streams respectively produce manipulation and authenticity evidence, which are progressively refined through cascaded multi-level fusion, bidirectional disagreement suppression, and edge-guided dynamic evidence debate. Based on the resulting competing evidence, an RL-based judge performs strategic re-inference and refinement on uncertain regions. The judge is optimized using an advantage-based policy optimization strategy with a soft-IoU objective, while prediction reliability is calibrated using entropy and cross-hypothesis consistency. In this way, CSAF extends existing RL-based localization from single-hypothesis sequential refinement to dual-hypothesis evidence reasoning and uncertainty-aware adjudication.

\section{Methodology}
We propose a courtroom-style adjudication framework (CSAF) comprising two stages, namely courtroom debate and judge's ruling, as illustrated in Fig.~\ref{fig:2}. In the courtroom debate stage, we construct a two-stream prosecution and defense architecture on top of a shared encoder via lightweight adapters,  and propose a dynamic debate mechanism to enhance the representation of evidence features. In the subsequent judge's ruling stage, the model aggregates evidence from multiple sources and employs a policy network based on RL to identify highly uncertain regions. Rather than treating all pixels equally, the judge performs adaptive conditional refinement on these disputed areas, ultimately producing the manipulation mask and the corresponding reliability scores.

\subsection{Courtroom Debate}
In the dual-hypothesis segmentation architecture, the prosecution and defense branches independently learn from the perspectives of manipulation and authenticity, which inevitably introduces information bias. To bridge this bias, we aim to enable both branches to view each other's evidence, allowing them to obtain additional information from different perspectives. However, in regions of disagreement, we need to prevent one branch from being misled by the evidence of the other. To address this issue, we propose a dynamic debate mechanism that adjusts the interaction between the prosecution and defense streams, allowing each branch to maintain higher distinguishability in its area of expertise while retaining independence in regions of disagreement. This effectively prevents the misguidance of information. Specifically, we introduce a divergence-suppression term into the bidirectional cross-attention module to impose principled constraints on contentious regions during feature fusion. We first compute a spatial disagreement map $\mathbf{D} \in \mathbb{R}^{1 \times H \times W}$ by calculating the mean squared difference over the channel dimension between the input features $\bm{mf}$ and $\bm{af} \in \mathbb{R}^{C \times H \times W}$:
\begin{equation}
  \mathbf{Df} = \frac{1}{C} \sum_{c=1}^{C} (\bm{mf}^{(c)} - \bm{af}^{(c)})^2.
\end{equation}
 We use $h$ heads with per-head dimension $d = C/h$. We apply $1\times1$ linear projections to $\bm{mf}$ and $\bm{af}$ to obtain $\mathbf{M}_Q^{(i)}$, $\mathbf{M}_K^{(i)}$, $\mathbf{M}_V^{(i)}$ and $\mathbf{A}_Q^{(i)}$, $\mathbf{A}_K^{(i)}$, $\mathbf{A}_V^{(i)}$, respectively. We then define the attention from the prosecution stream to the defense stream and vice versa as $\mathbf{SM}^{(i)}$ and $\mathbf{SA}^{(i)}$. This can be formulated as:
\begin{equation}
    \mathbf{SM}^{(i)} = \text{Softmax}\left(\frac{\mathbf{M}_Q^{(i)}(\mathbf{A}_K^{(i)})^\top}{\sqrt{d}} - \lambda \mathbf{Df}\right) \mathbf{A}_{V}^{(i)}
\end{equation}
where $\lambda$ denotes a suppression coefficient. The attention map $\mathbf{SA}^{(i)}$ is computed analogously. This formulation explicitly down-weights attention in regions with high conflict (large $\mathbf{Df}$), ensuring that each branch absorbs context only from reliable corresponding regions in the peer branch. The aggregated features are fused with original inputs via residual connections to obtain $\mathbf{MF}$ and $\mathbf{AF}$.

After cross-branch interaction, we adaptively reallocate evidence according to the response strength of each branch: at each spatial location, the features of the stronger branch are amplified, while those of the weaker branch are mildly suppressed, thereby enforcing a ``strong-get-stronger, weak-yield” debate pattern. We define a bounded difference map $\bm{\Delta}$ and a gating map $\bm{\alpha}$:
\begin{equation}
    \bm{\Delta} = \tanh(\mathbf{MF} - \mathbf{AF}), \quad \bm{\alpha} = \sigma(\text{Conv}([\mathbf{MF}, \mathbf{AF}])),
\end{equation}
where $\sigma(\cdot)$ is the sigmoid function and $[\cdot, \cdot]$ denotes concatenation. The features are updated via a symmetric push-pull operation:
\begin{equation}
{\hat{\mathbf{MF}}}= \mathbf{MF} + \bm{\alpha} \cdot \bm{\Delta}, \quad \hat{\mathbf{AF}}= \mathbf{AF} - \bm{\alpha} \cdot \bm{\Delta}
\end{equation}
Intuitively, when $\mathbf{MF}$ is stronger than $\mathbf{AF}$ at a given location, $\bm{\Delta}>0$ and the update amplifies $\hat{\mathbf{MF}}$ while suppressing $\hat{\mathbf{AF}}$ at that position. The opposite holds when $\mathbf{AF}$ is stronger. The gating factor $\bm{\alpha}$ is adaptively predicted from local features, so this pull–push update is applied only in regions with sufficient evidence and with a controlled adjustment magnitude. Meanwhile, the refinement preserves the total response of the two branches at each spatial location, $i.e.,$ $\hat{\mathbf{MF}} + \hat{\mathbf{AF}} = \mathbf{MF} + \mathbf{AF}$.
This shows that our method does not simply rescale the overall energy, but instead locally reallocates evidence between the prosecution and defense branches,  allowing the more reliable branch to dominate the representation at each spatial location.

To enhance the contrast and discriminability between the two types of evidence, we explicitly extract boundary cues from the input image, motivated by the fact that authentic and manipulated regions often share consistent geometric boundaries. Specifically, we apply a Laplacian operator to the source image $\mathbf{I}$ to capture high-frequency details and obtain the raw edge map $\mathbf{E}_{raw}$:
\begin{equation}
\mathbf{E}_{raw} = \text{ReLU}(\text{BN}(\text{Laplace}(\mathbf{I})))
\end{equation}
where BN denotes batch normalization. We then project $\mathbf{E}_{raw}$ into the feature space via a residual block and fuse it with multi-scale encoder features to inject semantic context. Taking $\hat{\mathbf{MF}}$ as an example, we concatenate the low-level backbone feature with the high-level feature $\hat{\mathbf{MF}}$ and apply a $1\times1$ convolution to obtain the contextual feature $\mathbf{F}_{ctx}$. Next, $\mathbf{F}_{ctx}$ is concatenated with $\mathbf{E}_{raw}$ and fed into another $1\times1$ convolution, followed by CBAM~\cite{woo2018cbam} to enhance informative boundaries along both channel and spatial dimensions. The final boundary prediction $\mathbf{tE}$ is formulated as:
\begin{equation}
\mathbf{tE} = \text{CBAM}(\text{Conv}([\mathbf{E}_{raw}, \mathbf{F}_{ctx}]))
\end{equation}
Finally, we adopt EFM~\cite{sun2022boundary} to inject the extracted boundary information $\mathbf{tE}$ into the feature $\hat{\mathbf{MF}}$, producing $\mathbf{tF}$. We then apply a $1\times1$ convolution to $\mathbf{tF}$ to squeeze the channel dimension to 1, producing the predicted mask $\mathbf{tP}$.

In summary, the prosecution branch outputs the manipulated-region feature $\mathbf{tF}$, the manipulated-region mask $\mathbf{tP}$, and the corresponding boundary map $\mathbf{tE}$. Meanwhile, the defense branch produces the authentic-region feature $\mathbf{rF}$, the authentic-region mask $\mathbf{rP}$, and the corresponding boundary map $\mathbf{rE}$.

\subsection{Judge’s Ruling}
\label{sec:judge}

Direct fusion of the prosecution prediction $\mathbf{tP}$ and defense prediction $\mathbf{rP}$ often lacks reliability due to the spatial inconsistency of forensic cues. To address this, we propose a judge model that employs an RL-based patch-level strategy to arbitrate between conflicting predictions. 

\textbf{Evidence Aggregation.}
The judge constructs a multi-source evidence feature by aggregating the prediction and edge masks from both branches with frequency-domain features extracted via Laplacian, SRM, and block-DCT filters. This tensor is processed by a lightweight convolutional encoder to form the initial evidence $\mathbf{V}$. To further integrate semantic context, we employ two sequential MLP-based adapters that project the high-level features from the prosecution branch $\mathbf{tF}$ and defense branch $\mathbf{rF}$ into the evidence space, yielding an enhanced representation $\mathbf{EV}$ for precise local adjudication:
\begin{equation}
   \mathbf{EV} =  \mathcal{A}_{t}(\mathcal{A}_{t}(\mathbf{V}+\mathbf{tF})+\mathbf{rF}),
\end{equation}
where $\mathcal{A}_{t}(\cdot)$ denotes the 2D MLP Adapter. Next, $\mathbf{EV}$ is processed through two $3 \times 3$ convolutions and a $1 \times 1$ convolution to generate a single-channel pixel-level dispute map $\mathbf{dM}$. Larger values in $\mathbf{dM}$ indicate pixels where the prosecution and defense branches exhibit strong disagreement.

\textbf{State Space Construction.}
The core mission of the judge model is to devise optimal processing strategies for local image regions. To facilitate fine-grained decision-making, we partition the input image and feature maps into $N$ non-overlapping patches. For each patch $i$, we formulate the judge's observation state as a 7-dimensional vector $\bm{s}_i$, constructed to comprehensively capture local characteristics regarding conflict and uncertainty:
\begin{equation}
   \bm{s}_{i}=[\mu_{i}, \sigma_{i}, \max\nolimits_{i}, \mathcal{H}_{i}, \bar{M}_{i}, \bar{D}_{i}, \bar{U}_{i}],
\end{equation}
where $[\mu_{i}, \sigma_{i}, \max_{i}, \mathcal{H}_{i}]$ correspond to the mean, standard deviation, maximum, and Shannon entropy of the evidence features $\mathbf{EV}$ within the patch. The remaining three components quantify divergence: $\bar{M}_i$ is the average dispute score derived from $\mathbf{dM}$. $\bar{D}_i$ represents the patch-wise mean of the absolute consistency gap $\mathbf{D} = |\mathbf{tP} - (\mathbf{1}-\mathbf{rP})|$, measuring the conflict between the prosecution's manipulation prediction and the defense's inverted authenticity prediction. $\bar{U}_i$ indicates the aggregated predictive uncertainty, computed via $\mathbf{U} = \mathcal{H}(\mathbf{tP}) + \mathcal{H}(\mathbf{1}-\mathbf{rP})$.

\textbf{Actor-Critic Decision Process.}
The judge operates within an Actor-Critic architecture. The Actor network $\pi_\theta$ maps the local state $\bm{s}_i$ to a discrete action space $|\mathcal{A}|=3$. These actions (conservative, correction, and reconstruction) function as conditional codes embedded into the subsequent segmentation network. To enable end-to-end differentiable sampling, we employ the Gumbel-Softmax technique with the straight-through estimator (STE). 
For state $\bm{s}_i$, $\pi_\theta$ outputs logits $\mathbf{z}_i$. We introduce stochasticity by adding Gumbel noise $\mathbf{g}_i \sim \text{Gumbel}(0,1)$ and computing the soft action vector ${y}_{i}^{soft}$:
\begin{equation}
y_{i, k}^{soft} = \frac{\exp((\mathbf{z}_{i, k} + \mathbf{g}_{i, k}) / \tau_g)}{\sum_{j=1}^{|\mathcal{A}|} \exp((\mathbf{z}_{i, j} + \mathbf{g}_{i, j}) / \tau_g)},
\end{equation}
where $\tau_g$ is the temperature parameter. We determine the discrete action index $a_i$ for the $i$-th patch via an argmax operation:
\begin{equation}
a_i = \underset{k}{\arg \max } (y_{i, k}^{ {soft }}), \quad {y}_{i}^{ {hard }} = {one\_hot}(a_i).
\end{equation}
where $\underset{k}{\arg\max(\cdot)}$ denotes the operation of retrieving the index of the maximum value, and ${one\_hot}(\cdot)$ denotes the one-hot encoding operation that converts this index into a binary vector. To allow backpropagation through the sampling process, the final action vector $\bm{Ac}_i$ is formulated using STE:
\begin{equation}
\mathbf{Ac}_i = \text{sg}({y}_{i}^{hard} - {y}_{i}^{soft}) + {y}_{i}^{soft},
\end{equation}
where $\text{sg}(\cdot)$ denotes the stop-gradient operator. Finally, the collection of action vectors $\{ \mathbf{Ac}_i \}_{i=1}^N$ is spatially reshaped to form the action map $\mathbf{Ac}$. This map, concatenated with the evidence features and state statistics, is fed into the lightweight U-shaped segmentation network to derive the final verdict $\mathbf{PM}$.

\begin{table}[!t]
\centering
\caption{The datasets used in our experiments. CM, SP, and IP indicate three common image manipulation types: copy-move, splicing, and inpainting.}
\label{tab:datasets}
\resizebox{0.47\textwidth}{!}{%
\begin{tabular}{@{}ccccccc@{}}
\toprule
\textbf{Dataset} & \textbf{Nums} & \textbf{\#CM} & \textbf{\#SP} & \textbf{\#IP} & \textbf{Train} & \textbf{Test} \\ 
\midrule
CASIAv2~\cite{dong2013casia} & 5123 & 3295 & 1828 & 0 & 5123 & 0 \\ 
Coverage~\cite{wen2016coverage} & 100 & 100 & 0 & 0 & 70 & 30 \\ 
NIST16~\cite{guan2019mfc} & 564 & 68 & 288 & 208 & 383 & 181 \\ 
CASIAv1~\cite{dong2013casia} & 920 & 459 & 461 & 0 & 0 & 920 \\
Columbia~\cite{hsu2006columbia} & 180 & 0 & 180 & 0 & 0 & 180 \\ 
DSO~\cite{de2013exposing} & 100 & 0 & 100 & 0 & 0 & 100 \\
IMD2020~\cite{Novozamsky_2020_WACV} & 2010 & -- & -- & -- & 0 & 2010 \\
CocoGlide~\cite{nichol2021glide} & 512 & - & -& - & 0 &512 \\
ITW~\cite{huh2018fighting} & 201 & 0 & 201 & 0 &0 &201\\
\bottomrule
\end{tabular}%
}
\end{table}

\textbf{Reinforcement Learning Objective.}
To optimize the policy, we employ a relative gain strategy that incentivizes the judge to intervene only when its re-reasoning yields a tangible improvement over the raw evidence.
First, we construct a strong heuristic baseline $\mathbf{B} = \max(\mathbf{tP}, \mathbf{1}-\mathbf{rP})$, which represents the optimal deterministic outcome achievable by simply accepting the most confident cues from either branch without complex arbitration.
Consequently, the reward $r$ is formulated as the relative improvement in Soft-IoU:
\begin{equation}
    r = \mathcal{J}_{iou}(\mathbf{PM}, \mathbf{G}) - \mathcal{J}_{iou}(\mathbf{B}, \mathbf{G}),
\end{equation}
where $\mathcal{J}_{iou}$ denotes the Soft-IoU metric. The judge model is optimized within an Actor-Critic framework. The Actor $\pi_\theta$ is updated via the policy gradient to maximize the expected relative gain, while the Critic $V_\phi$ learns to estimate this gain to further reduce gradient variance. The joint objective functions are defined as:
\begin{align}
    \mathcal{L}_{pg} &= -\frac{1}{N}\sum_{i=1}^{N}
\operatorname{sg}\!\left(r-V_\phi(\mathbf{s}_i)\right)
\log\pi_\theta(a_i\mid\mathbf{s}_i). \\
    \mathcal{L}_{val} &= \frac{1}{N} \sum_{i=1}^N \| V_\phi(\bm{s}_i) - \text{sg}(r) \|^2,
\end{align}

Minimizing $\mathcal{L}_{pg}$ is equivalent to performing gradient ascent on the expected reward, directing the Actor $\pi_\theta$ to increase the probability of actions that yield positive relative gains. Simultaneously, by minimizing $\mathcal{L}_{val}$, the Critic $V_\phi$ is trained to regress the relative gain signal, serving as an auxiliary stabilizer that encourages consistent policy evaluation and improves training stability.

\textbf{Reliability-Aware Consistency and Calibration.}
Although the judge performs policy-driven arbitration, dual-hypothesis learning may still degenerate into trivial agreement in easy regions or become overconfident under noisy evidence. To alleviate these issues, we introduce a reliability-aware consistency regularization. Specifically, the judge predicts a pixel-wise reliability map $\mathbf{Rel}$ from the evidence representation $\mathbf{EV}$, which is used to gate the consistency constraint so that agreement is enforced only on trustworthy pixels. We encourage the prosecution prediction $\mathbf{tP}$ and the complementary defense prediction $(\mathbf{1}-\mathbf{rP})$ to be consistent in reliable, non-boundary regions by minimizing the symmetric KL divergence (SymKL):
\begin{equation}
\mathbf{M}_{gate} = \mathbb{1}(\mathbf{Rel} > \tau)\times(1-\mathbf{tE})\times(1-\mathbf{rE}),
\end{equation}
\begin{equation}
\mathcal{L}_{c} = \frac{\sum \mathbf{M}_{gate} \cdot \text{SymKL}(\mathbf{tP}\ ||\ (\mathbf{1}-\mathbf{rP}))}{\sum \mathbf{M}_{gate}+\epsilon}
\end{equation}
where $\tau=0.6$. By excluding unreliable or boundary-ambiguous pixels, $\mathcal{L}_{c}$ prevents mode collapse while avoiding forced agreement on genuinely uncertain regions. Crucially, the effectiveness of $\mathcal{L}_{c}$ hinges on the quality of $\mathbf{Rel}$. To ensure $\mathbf{Rel}$ accurately reflects prediction confidence, we impose a calibration objective $\mathcal{L}_{cal}$ using a pseudo-label $\mathbf{R}^*$ constructed from prediction entropy and inter-branch agreement:
\begin{equation}
\mathcal{L}_{cal} = \mathcal{L}_{bce}(\mathbf{Rel}, \mathbf{R}^*) + \beta \cdot ||\mathbf{PM} - \mathbf{G}||_2^2
\end{equation}
where $\mathbf{G}$ denotes the ground truth, $\beta=0.1$, and $\mathbf{R}^* = 1 - 0.5 \cdot \text{Norm}(\mathcal{H}(\mathbf{PM})) - 0.5 \cdot \text{Norm}(|\mathbf{tP} - (\mathbf{1}-\mathbf{rP})|).$ In essence, $\mathcal{L}_{cal}$ achieves probability calibration by encouraging $\mathbf{Rel}$ to exhibit low-entropy and high-consistency patterns, while penalizing overconfident predictions. In summary, we define the reliability loss $\mathcal{L}_{rel}$ as follows:
\begin{equation}
    \mathcal{L}_{rel} = \mathcal{L}_{cal} + \lambda_{c}\mathcal{L}_{c}
\end{equation}
In our experiments, we set $\lambda_{c}=0.1$.
\subsection{Loss Function}

For the prosecution prediction map $\mathbf{tP}$, the defense prediction map $\mathbf{rP}$, and the final verdict $\mathbf{PM}$ produced by the judge model, we impose a structure-consistency loss $\mathcal{L}_{s}$~\cite{wei2020f3net} to emphasize hard-to-handle pixels, thereby improving the accuracy of IML. For the boundary predictions $\mathbf{tE}$ and $\mathbf{rE}$, considering the severe class imbalance between edge and non-edge samples, we adopt an edge loss $\mathcal{L}_{e}$ (the sum of BCE and Dice losses) to enforce boundary alignment and enhance edge discriminability.
\begin{equation}
    \mathcal{L}_{seg} = \mathcal{L}_{s}(\mathbf{tP}, \mathbf{G}) + \mathcal{L}_{s}(\mathbf{rP}, \mathbf{1-G}) + \mathcal{L}_{s}(\mathbf{PM}, \mathbf{G})
\end{equation}
\begin{equation}
    \mathcal{L}_{bg} = \mathcal{L}_{e}(\mathbf{tE}, \mathbf{G_e}) +  \mathcal{L}_{e}(\mathbf{rE}, \mathbf{G_e})
\end{equation}
where $\bm{G}_e$ denotes the edge ground truth. Note that $\mathbf{rP}$ is supervised by $\mathbf{1-G}$ to predict authentic regions. The overall loss function is defined as:
\begin{equation}
    \mathcal{L}_{all} = \mathcal{L}_{seg} + \mathcal{L}_{bg} + \mathcal{L}_{rel} + \lambda_{rl}(\mathcal{L}_{pg} + \mathcal{L}_{val})
\end{equation}
In our experiments, we set $\lambda_{rl}= 0.1$.

\begin{table*}[!t]
\centering
\caption{Performance comparison with other advanced IML methods in terms of
F1 score (fixed threshold at 0.5). The \textbf{best} and
\underline{second-best} results are marked in bold and underlined,
respectively.}
\label{tab:main_results}

\resizebox{\textwidth}{!}{%
\begin{tabular}{lcccccccccccc}
\toprule
\multirow{2}{*}{\textbf{Method}} &
\multirow{2}{*}{\textbf{Pub.}} &
\multicolumn{4}{c}{\textbf{In-Distribution (ID)}} &&
\multicolumn{6}{c}{\textbf{Out-of-Distribution (OOD)}} \\
\cmidrule(lr){3-6}
\cmidrule(lr){8-13}
& & CASIAv1 & Coverage & NIST16 & Avg. &&
Columbia & DSO & CocoGlide & ITW & IMD2020 & Avg. \\
\midrule

PSCC-Net~\cite{liu2022pscc}
& TCSVT'22
& 0.358 & 0.320 & 0.302 & 0.326
&& 0.572 & 0.322 & 0.328 & 0.362 & 0.272 & 0.371 \\

MFI-Net~\cite{ren2023mfi}
& TCSVT'24
& 0.571 & 0.448 & 0.432 & 0.483
&& 0.342 & 0.081 & 0.204 & 0.255 & 0.331 & 0.243 \\

PIM-Net~\cite{bai2025pim}
& PR'25
& 0.587 & \underline{0.655} & 0.496 & 0.579
&& 0.669 & 0.117 & 0.324 & 0.314 & 0.372 & 0.359 \\

PIM~\cite{kong2025pixel}
& TPAMI'25
& 0.577 & 0.555 & 0.411 & 0.514
&& 0.626 & 0.165 & 0.269 & 0.332 & 0.345 & 0.347 \\

Mesorch~\cite{zhu2025mesoscopic}
& AAAI'25
& 0.649 & 0.449 & 0.440 & 0.512
&& 0.655 & 0.196 & 0.159 & 0.294 & 0.364 & 0.334 \\

BPIML~\cite{guo2025passive} & AAAI'26 & 0.543 & 0.418 & 0.409 & 0.457 && 0.378 & 0.253 & 0.237 & 0.305 & 0.321 & 0.299  \\

RelayFormer~\cite{huang2026relayformer} & ICLR'26 & 0.567 & 0.324 & 0.372 &0.421 &&  0.507 & 0.113 & 0.239 & 0.277 & 0.311 &0.289\\

GFrame~\cite{yu20262d}
& ACM MM'26
& 0.627 & 0.592 & 0.556 & 0.592
&& \underline{0.797}
& 0.205
& \underline{0.381}
& \pmb{0.450}
& \pmb{0.528}
& \underline{0.472} \\

PNF-Net~\cite{li2026prototype}
& TCSVT'26
& \underline{0.694}
& 0.649
& \underline{0.557}
& \underline{0.633}
&& 0.757
& \underline{0.334}
& 0.366
& 0.377
& 0.454
& 0.458 \\

\midrule

Ours
& --
& \pmb{0.734}
& \pmb{0.722}
& \pmb{0.593}
& \pmb{0.683}
&& \pmb{0.811}
& \pmb{0.342}
& \pmb{0.408}
& \underline{0.428}
& \underline{0.464}
& \pmb{0.491} \\

\bottomrule
\end{tabular}%
}
\end{table*}

\begin{figure*}[!t]
  \centering
   \includegraphics[width=\linewidth]{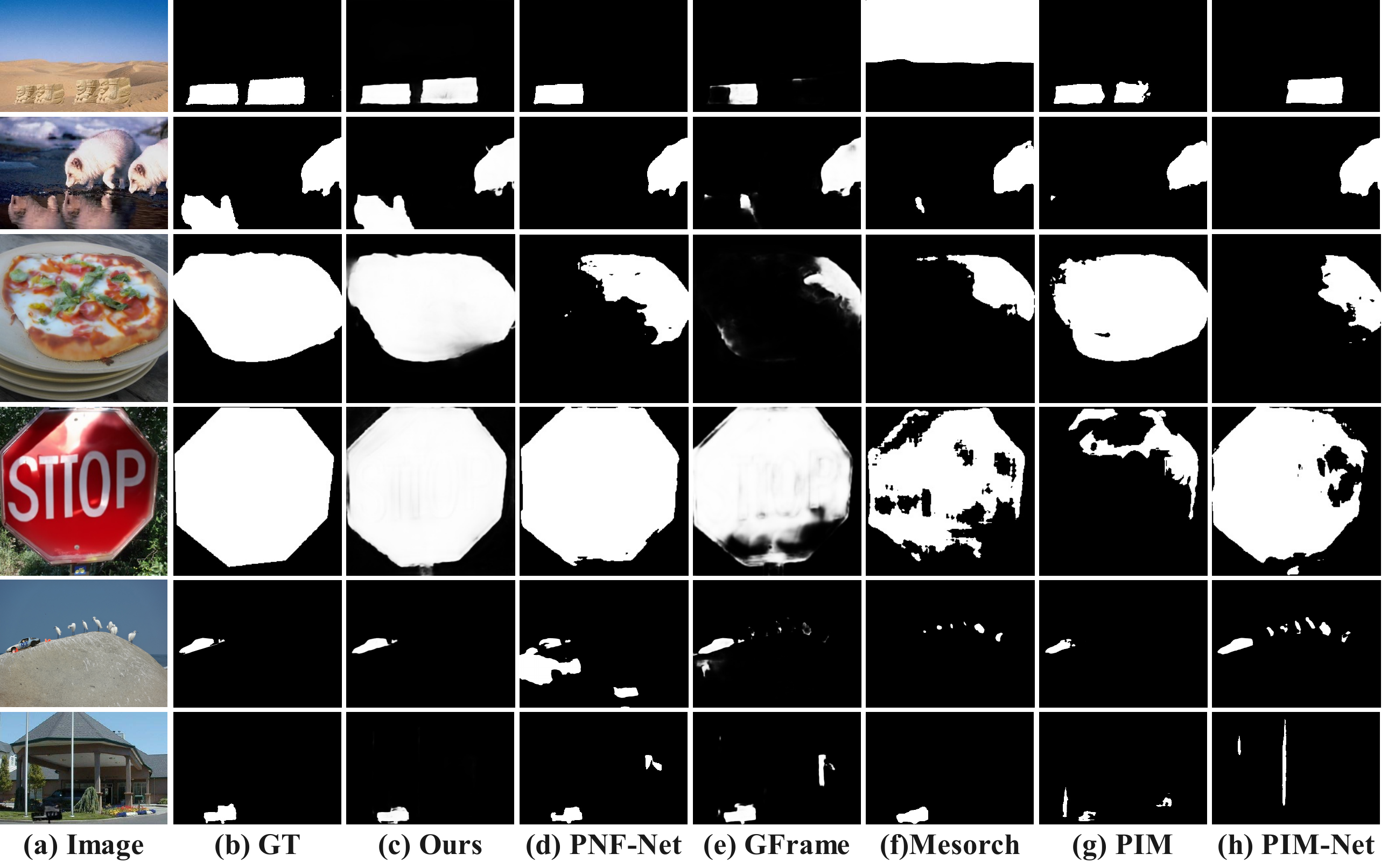}

   \caption{Visualization comparison of our method compared with other SOTA IML methods. GT represents ground truth.}
   \label{fig:vs}
\end{figure*}

\section{Experiments and results}

\subsection{Datasets and Implementation Details}
The datasets used in our experiments and their corresponding splits are summarized in Table~\ref{tab:datasets}. For NIST16~\cite{guan2019mfc}, we follow the data processing protocol of Ma et al.~\cite{ma2025imdl}. We adopt PVTv2~\cite{wang2022pvt} initialized with pretrained weights as the backbone. All RGB images, manipulation masks, and boundary masks are resized to $416\times416$ pixels. Bilinear interpolation with anti-aliasing is applied to RGB images, while nearest-neighbor interpolation is used for both types of masks to preserve their discrete labels. The RGB inputs are normalized using the ImageNet mean and standard deviation. The entire network, including the dual-branch localization model and the judge, is trained end-to-end for 20 epochs using AdamW~\cite{loshchilov2018decoupled}. The total batch size is set to 12, with an initial learning rate of $1\times10^{-4}$ and a weight decay of $1\times10^{-4}$. The learning rate is reduced by a factor of 10 at the 15th epoch. Training is conducted on two NVIDIA RTX 3090 Ti GPUs. For fair comparison with existing methods, we report the F1 score using a fixed threshold of 0.5 as the evaluation metric.


\begin{table*}[!t]
\centering
\caption{Ablation study of the proposed modules in terms of F1 score
(fixed threshold at 0.5). The \textbf{best} results are marked in bold.}
\label{ab}

\resizebox{\textwidth}{!}{%
\begin{tabular}{lccccccccccc}
\toprule
\multirow{2}{*}{\textbf{Method}} &
\multicolumn{4}{c}{\textbf{In-Distribution (ID)}} &&
\multicolumn{6}{c}{\textbf{Out-of-Distribution (OOD)}} \\
\cmidrule(lr){2-5}
\cmidrule(lr){7-12}
& CASIAv1 & Coverage & NIST16 & Avg. &&
Columbia & DSO & CocoGlide & ITW & IMD2020 & Avg. \\
\midrule

Ours \textit{w/o} Debate
& 0.675 & 0.603 & 0.514 & 0.598
&& 0.708 & 0.294 & 0.257 & 0.408 & 0.468 & 0.427 \\

Ours \textit{w/o} Judge Model
& 0.663 & 0.658 & 0.500 & 0.607
&& 0.562 & 0.291 & 0.289 & 0.385 & 0.441 & 0.394 \\

Judge Model \textit{w/o} RL
& 0.662 & 0.695 & 0.559 & 0.639
&& 0.646 & 0.321 & 0.263 & 0.419 & 0.461 & 0.422 \\

Ours \textit{w/o} Reliability Loss
& 0.661 & 0.689 & 0.517 & 0.623
&& 0.705 & \textbf{0.378} & 0.311 & \textbf{0.418}
& \textbf{0.463} & 0.455 \\

\midrule

Ours
& \textbf{0.734}
& \textbf{0.722}
& \textbf{0.593}
& \textbf{0.683}
&& \textbf{0.811}
& 0.342
& \textbf{0.408}
& 0.428
& 0.464
& \textbf{0.491} \\

\bottomrule
\end{tabular}
}
\end{table*}

\subsection{Comparison with SOTA Methods}
Since the source code of MFI-Net has not been publicly released, we reimplemented the method based on the implementation details provided in the original paper. For all other methods, we used the official source code released by their respective authors. To ensure a fair comparison, all methods were trained for 20 epochs using the same data preprocessing pipeline, identical training and test splits, and a consistent experimental environment.

\textbf{Image manipulation localization.} Table~~\ref{tab:main_results} compares our method with recent state-of-the-art image manipulation localization (IML) approaches under a fixed threshold of 0.5. Our method achieves the best average performance on both the in-distribution (ID) and out-of-distribution (OOD) benchmarks. The performance gains in the ID setting are closely associated with the explicit modeling of two competing hypotheses. Rather than treating authentic-region supervision merely as an auxiliary constraint, we employ a prosecution stream and a defense stream to model manipulated and authentic regions, respectively, based on representations extracted by a shared multi-scale encoder. The two streams first aggregate hierarchical features through cascaded multi-scale fusion and subsequently exchange information via bidirectional cross-attention. Disagreement suppression and dynamic debate refinement are further introduced to adaptively adjust the feature representations in conflicting regions, thereby preventing either hypothesis from dominating the prediction in ambiguous areas. In addition, the edge branches integrate a fixed Laplacian prior with visual features at different levels and use the resulting boundary information to modulate both hypothesis streams. This design enhances the sensitivity to manipulation boundaries while reducing false positives caused by visually salient yet authentic structures. The resulting region- and boundary-level evidence provides complementary information for the subsequent adjudication process, which is consistent with the leading performance observed on CASIAv1~\cite{dong2013casia}, Coverage~\cite{wen2016coverage}, and NIST16~\cite{guan2019mfc}.

Under OOD settings, the judge does not directly adopt the output of either hypothesis stream. Instead, it jointly considers the prosecution probability, the complementary defense probability, the two edge responses, and intermediate features generated during the debate process. The judge further incorporates image residual and frequency-domain forensic cues, including Laplacian, SRM, and block-DCT responses. Compared with high-level semantic representations, these low-level forensic cues are generally less dependent on specific image content and semantic categories, and therefore provide additional manipulation evidence under unseen data distributions. Based on cross-hypothesis discrepancy, prediction entropy, dispute intensity, and local evidence statistics, the policy module selects an appropriate adjudication action for each local image region. Its advantage-based objective uses the soft-IoU improvement of the current adjudication result over the initial dual-hypothesis baseline as the reward signal. This encourages the policy to learn region-specific adjudication strategies that improve the overall localization quality, rather than applying a fixed and uniform fusion rule to regions with different levels of disagreement. We further introduce a reliability calibration mechanism that jointly considers prediction entropy and cross-hypothesis discrepancy to construct pixel-wise reliability estimates. The complementary constraint, jointly gated by reliability and boundary information, is primarily imposed on reliable non-boundary regions, encouraging consistency between the prosecution prediction and the complementary defense prediction in unambiguous areas while avoiding overly strong constraints around intrinsically uncertain boundaries. This design preserves the adversarial and complementary nature of the two hypotheses while reducing the influence of unreliable evidence during optimization. Its effectiveness is consistent with the leading results achieved on Columbia~\cite{hsu2006columbia}, DSO~\cite{de2013exposing}, and CocoGlide~\cite{nichol2021glide}. Although GFrame~\cite{yu20262d} obtains higher F1 scores on ITW~\cite{huh2018fighting} and IMD2020~\cite{Novozamsky_2020_WACV}, our method still achieves the best overall average performance across the OOD benchmarks. These results suggest that competitive hypothesis interaction and reliability-aware adjudication provide a favorable balance between in-distribution localization accuracy and cross-dataset generalization.







\textbf{Visual comparison.} As shown in Fig.~\ref{fig:vs}, to further qualitatively validate the effectiveness of our model, we conducted visual comparisons across ID and OOD settings, targeting three challenging scenarios: multi-object, large-object, and small-object manipulation. Thanks to the reinforcement of semantic consistency by the dynamic debate mechanism, our model successfully resolves the internal void issue in large-object masks and eliminates boundary adhesion between multiple instances. Furthermore, the RL-driven judge model effectively disentangles semantic interference and suppresses uncertainty in ambiguous regions. This capability enables the model to not only precisely localize small-scale manipulation targets but also sharply delineate their fine-grained edges.

\begin{figure*}[!t]
  \centering
   \includegraphics[width=\linewidth]{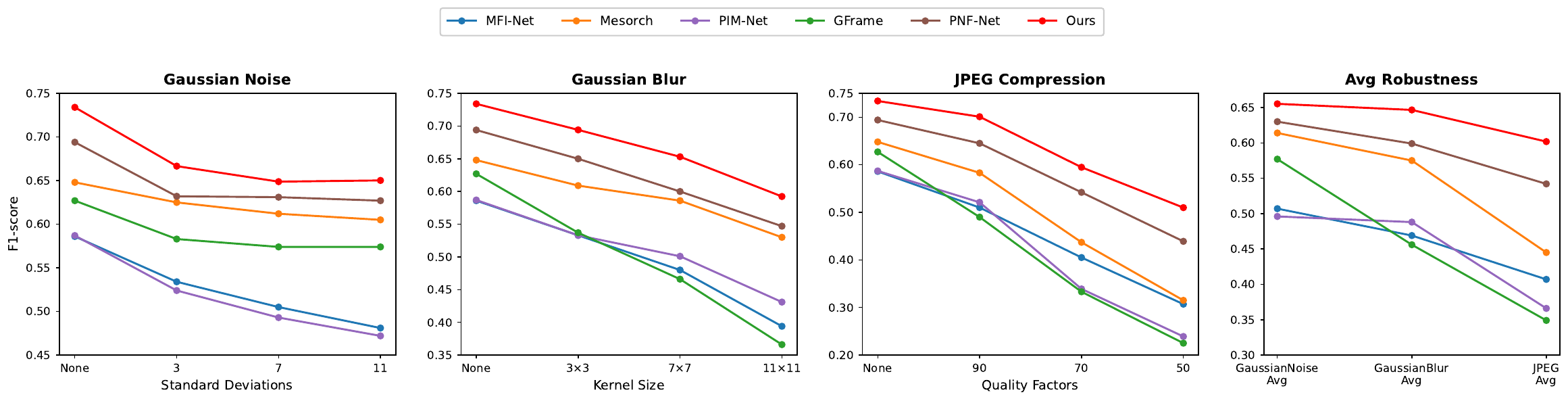}

   \caption{Robustness analysis under standard perturbations on the CASIAv1 dataset.}
   \label{fig:4}
\end{figure*}

\begin{table}[t]
\centering
\caption{Performance on images processed by online social media platforms.}
\label{tab:osn}

\resizebox{0.47\textwidth}{!}{%
\begin{tabular}{lccccc}
\toprule
\multirow{2}{*}{\textbf{Method}} &
\multicolumn{5}{c}{\textbf{Online Social Media Compression (F1-score)}} \\
\cmidrule(lr){2-6}
& Facebook & WeChat & Weibo & WhatsApp & \textbf{Avg.} \\
\midrule

MFI-Net
& 0.531 & 0.468 & 0.530 & 0.533 & 0.516 \\

Mesorch
& 0.600 & 0.488 & 0.604 & 0.612 & 0.576 \\

PIM-Net
& 0.611 & 0.524 & 0.564 & 0.621 & 0.580 \\

GFrame
& 0.569 & 0.459 & 0.575 & 0.581 & 0.546 \\

PNF-Net
& 0.627 & 0.483 & 0.620 & 0.633 & 0.591 \\

\midrule

Ours
& \pmb{0.710}
& \pmb{0.608}
& \pmb{0.647}
& \pmb{0.710}
& \pmb{0.669} \\

\bottomrule
\end{tabular}%
}
\end{table}

\subsection{Ablation Study}


\textbf{Effectiveness of the Dynamic Debate Mechanism:} As shown in Table~\ref{ab}(a), removing the dynamic debate mechanism degrades performance primarily because the model loses its ability to resolve feature conflicts through interaction. Without this mechanism, the model cannot penalize semantic inconsistencies between the prosecution and defense streams, so noisy features in ambiguous regions are not effectively suppressed. Moreover, the absence of bidirectional feature correction prevents the model from exploiting the adversarial push–pull dynamics to sharpen decision boundaries and to compensate for semantic gaps.

\textbf{Effectiveness of the Judge Model:} As shown in Table~\ref{ab}(b), removing the judge model leads to a substantial performance drop, primarily because the model loses its capability for multimodal evidence fusion and uncertainty-aware error correction. Moreover, without the guidance of an IoU-based advantage reward, the model can no longer trigger strategic re-inference in high-entropy regions, thereby forfeiting a critical mechanism for targeted refinement of hard samples and for calibrating predictive confidence.

\textbf{Effectiveness of RL:} As shown in Table~\ref{ab}(c), removing RL causes the model to lose its capability for strategic re-inference. Under the actor–critic framework, RL leverages local statistics to adaptively select actions for each image patch, enabling targeted correction in regions with high uncertainty. More importantly, without RL, the model can no longer effectively exploit the IoU-based advantage signal via policy gradients to guide discrete action decisions. Consequently, in difficult regions with ambiguous boundaries or conflicting evidence, the model lacks the incentive to explore and refine near-optimal decision trajectories.

\textbf{Effectiveness of the Reliability Loss:} As shown in Table~\ref{ab}(d), the performance degradation observed after removing the reliability loss primarily results from the model’s loss of uncertainty calibration. This loss function achieves confidence calibration by encouraging the model to reduce its predictive confidence in logically inconsistent or high-entropy regions. Without this mechanism, the model is prone to overconfidence on ambiguous boundaries or hard samples and cannot effectively suppress low-quality predictions arising from conflicting decisions between the defense and prosecution, thereby substantially reducing the reliability of the final verdict.


\subsection{Robustness Evaluation}
To further demonstrate the strong robustness of our courtroom-style paradigm to post-processing, we evaluate the model under two representative degradation settings: (i) compression artifacts introduced by social-media transmission and (ii) common image corruptions. Specifically, following the evaluation protocol of MVSS-Net~\cite{dong2022mvss}, we test images compressed by Facebook, Weibo, WeChat, and WhatsApp. As reported in Table~\ref{tab:osn}, our method remains highly robust in real-world online sharing scenarios. In addition, Fig.~\ref{fig:4} summarizes the results under standard image degradations, including Gaussian noise, Gaussian blur, and JPEG compression, where our approach again exhibits exceptional robustness. These findings indicate that, compared with conventional trace-seeking methods that rely on fragile low-level artifacts, our courtroom-style framework achieves more reliable and stable IML under post-processing and external noise by enabling evidence-driven dual-hypothesis confrontation and uncertainty-aware adjudication with calibrated confidence through strategic re-inference.

\begin{table}[!t]
\caption{Ablation of the reinforcement learning weight $\lambda_{rl}$.}
\label{tab:rl}
\centering
\resizebox{0.47\textwidth}{!}{%
\begin{tabular}{@{}lccccc@{}}
\toprule
\multirow{2}{*}{\textbf{Distribution}} &
\multicolumn{5}{c}{\textbf{Hyperparameters}} \\
\cmidrule(lr){2-6}
 & $\lambda_{rl}=0.01$
 & $\lambda_{rl}=0.05$
 & $\lambda_{rl}=0.1$
 & $\lambda_{rl}=0.5$
 & $\lambda_{rl}=1$ \\
\midrule
\multicolumn{1}{c}{$\mathbf{Avg.ID}$}
 & 0.610 & 0.633 & \pmb{0.683} & 0.592 & 0.583 \\
\multicolumn{1}{c}{$\mathbf{Avg.OOD}$}
 & 0.425 & 0.478 & \pmb{0.491} & 0.425 & 0.427 \\
\bottomrule
\end{tabular}%
}
\end{table}

\subsection{Impact of Hyperparameter $\lambda_{rl}$}
Table \ref{tab:rl} presents the sensitivity analysis of the reinforcement learning loss weight, $\lambda_{rl}$, across both ID and OOD benchmarks. Overall, we observe a distinct ``optimal interval'' around $\lambda_{rl}=0.1$, which strikes the best balance between fitting ID data and improving OOD generalization. Crucially, $\lambda_{rl}=0.1$ yields the best average performance on both ID and OOD benchmarks, indicating that an appropriate RL weight improves not only in-domain localization accuracy but also cross-dataset generalization. This broad consistency indicates that an appropriate RL weight genuinely enhances robustness against various distribution shifts, rather than merely overfitting to a specific OOD scenario. From a mechanism perspective, the results can be interpreted as follows:

\textbf{Insufficient Incentive ($\lambda_{rl} \leq 0.05$):} When $\lambda_{rl}$ is too small, the RL term provides insufficient optimization signal to learn effective patch-level decisions. Consequently, the model behaves similarly to a purely supervised fusion framework: although ID performance remains acceptable, the cross-domain corrective effect of RL is not fully utilized, resulting in limited OOD robustness. Notably, at $\lambda_{rl}=0.05$, ID performance improves while OOD performance drops, suggesting a tendency toward overfitting.
    
\textbf{Gradient Instability ($\lambda_{rl} \geq 0.5$):} Conversely, when $\lambda_{rl}$ is too large, the inherent high variance of policy gradients is amplified and dominates the optimization landscape. This introduces instability and exploration noise that disrupts the stable convergence of the shared feature encoder, leading to consistent degradation in both ID and OOD performance.

\subsection{Impact of Hyperparameter $\lambda_{c}$}
The parameter $\lambda_{c}$ controls the strength of the complementary constraint between the prosecution and defense streams. Since the two streams represent competing hypotheses---evidence supporting manipulation and evidence supporting authenticity, respectively---an appropriate complementary constraint is expected to encourage hypothesis-specific predictions while preventing excessive redundancy between the two branches.

\begin{table}[!t]
\caption{Ablation of the complementary loss weight $\lambda_{c}$.}
\label{tab:lambda_comp}
\centering
\resizebox{0.47\textwidth}{!}{%
\begin{tabular}{@{}lccccc@{}}
\toprule
\multirow{2}{*}{\textbf{Distribution}} &
\multicolumn{5}{c}{\textbf{Hyperparameters}} \\
\cmidrule(lr){2-6}
 & $\lambda_{c}=0.01$
 & $\lambda_{c}=0.05$
 & $\lambda_{c}=0.1$
 & $\lambda_{c}=0.5$
 & $\lambda_{c}=1$ \\
\midrule
\multicolumn{1}{c}{$\mathbf{Avg.ID}$}
 & 0.638 & 0.646 & \pmb{0.683} & 0.614 & 0.610 \\
\multicolumn{1}{c}{$\mathbf{Avg.OOD}$}
 & 0.441 & 0.448 & \pmb{0.491} & 0.472 & 0.452 \\
\bottomrule
\end{tabular}%
}
\end{table}

\begin{table}[!t]
\caption{Ablation of the judge patch size.}
\label{tab:judge_patch}
\centering
\resizebox{0.47\textwidth}{!}{%
\begin{tabular}{@{}lccccc@{}}
\toprule
\multirow{2}{*}{\textbf{Distribution}} &
\multicolumn{5}{c}{\textbf{Hyperparameters}} \\
\cmidrule(lr){2-6}
 & $\mathrm{Patch}=4$
 & $\mathrm{Patch}=8$
 & $\mathrm{Patch}=16$
 & $\mathrm{Patch}=32$
 & $\mathrm{Patch}=64$ \\
\midrule
\multicolumn{1}{c}{$\mathbf{Avg.ID}$}
 & 0.517 & 0.623 & \pmb{0.683} & 0.614 & 0.590 \\
\multicolumn{1}{c}{$\mathbf{Avg.OOD}$}
 & 0.376 & 0.433 & \pmb{0.491} & 0.430 & 0.417 \\
\bottomrule
\end{tabular}%
}
\end{table}

As reported in Table~\ref{tab:lambda_comp}, increasing $\lambda_{c}$ from 0.01 to 0.1 improves the average ID F1 score from 0.638 to 0.683 and the average OOD F1 score from 0.441 to 0.491, corresponding to absolute gains of 4.5 and 5.0 percentage points, respectively. When $\lambda_{c}$ is too small, the interaction between the two streams is only weakly regularized, which may lead to redundant or insufficiently discriminative evidence. In this case, the judge receives less informative cues for comparing the competing hypotheses, particularly in ambiguous regions. The best overall performance is obtained at $\lambda_{c}=0.1$. This result suggests that a moderate complementary constraint provides a suitable balance between preserving stream-specific forensic evidence and maintaining meaningful disagreement between the two predictions. Such disagreement can provide useful information for uncertainty estimation and subsequent refinement. The simultaneous improvement on both ID and OOD datasets further suggests that this setting not only benefits in-domain localization but also improves robustness to distribution shifts. Further increasing $\lambda_{c}$ degrades the performance. At $\lambda_{c}=0.5$, the average ID and OOD F1 scores decrease to 0.614 and 0.472, respectively, and they further drop to 0.610 and 0.452 when $\lambda_{c}=1$. An excessively large complementary weight may overemphasize the separation between the two hypotheses, forcing their predictions to remain unnecessarily distinct even when the underlying evidence is ambiguous. This may suppress useful stream-specific forensic cues and interfere with the primary localization objective. Therefore, we set $\lambda_{c}=0.1$ in the remaining experiments.

\subsection{Impact of Judge Patch Size}
The patch size determines the spatial granularity at which the judge aggregates prosecution--defense evidence, their disagreement, and the associated uncertainty. It therefore controls the tradeoff between preserving subtle local forensic traces and providing sufficient spatial context for reliable patch-level adjudication.

As shown in Table~\ref{tab:judge_patch}, increasing the patch size from 4 to 16 improves the average ID F1 score from 0.517 to 0.683 and the average OOD F1 score from 0.376 to 0.491, corresponding to absolute gains of 16.6 and 11.5 percentage points, respectively. With a patch size of 4, each decision region contains only limited spatial context and may capture fragmented manipulation traces. Consequently, the aggregated evidence can be sensitive to local noise, compression artifacts, and minor prediction fluctuations. Although increasing the patch size to 8 improves both ID and OOD performance, the available context remains less effective than that obtained with a patch size of 16. A patch size of 16 achieves the best performance on both ID and OOD datasets. At this scale, the judge can incorporate sufficient contextual information to compare the competing hypotheses while retaining adequate spatial resolution for subtle artifacts and manipulation boundaries. This balance is particularly beneficial for patch-level uncertainty estimation and enables the refinement policy to focus on genuinely ambiguous regions rather than isolated noisy responses. The improved OOD performance also suggests that an appropriate adjudication scale reduces sensitivity to dataset-specific disturbances and facilitates the use of more robust forensic evidence. When the patch size is further increased to 32 and 64, the average ID F1 score decreases to 0.614 and 0.590, while the average OOD F1 score drops to 0.430 and 0.417, respectively. Large patches are more likely to contain heterogeneous content, including manipulated regions, authentic regions, and semantic boundaries. Aggregating such mixed responses can dilute weak manipulation traces and reduce the spatial precision of the disagreement and uncertainty estimates. As a result, the judge may assign the same refinement decision to regions with different forensic characteristics, leading to consistent performance degradation on both ID and OOD datasets.

\section{Conclusion}
In this work, we reformulate IML as a process of evidence confrontation followed by judgment, and propose an interactive closed-loop framework composed of prosecution, defense, and judge modules, thereby addressing the limitations of existing IML methods in the explicit modeling of authenticity evidence and adversarial localization reasoning. By explicitly modeling both manipulation and authenticity evidence, leveraging contrastive analysis to reveal regions of evidential conflict and divergence, and performing adaptive re-inference on uncertain regions, the proposed method achieves more precise localization and stronger robustness under cross-domain and degraded conditions. Experimental results demonstrate that this evidence-driven adversarial reasoning paradigm holds significant potential for the development of more generalizable visual forensic systems. Future work will further extend this framework to more complex forgery scenarios.

\ifCLASSOPTIONcaptionsoff
  \newpage
\fi
\bibliographystyle{IEEEtran}

\bibliography{reference}

\begin{IEEEbiography}[{\includegraphics[width=1.0in,height=1.2in,clip,keepaspectratio]{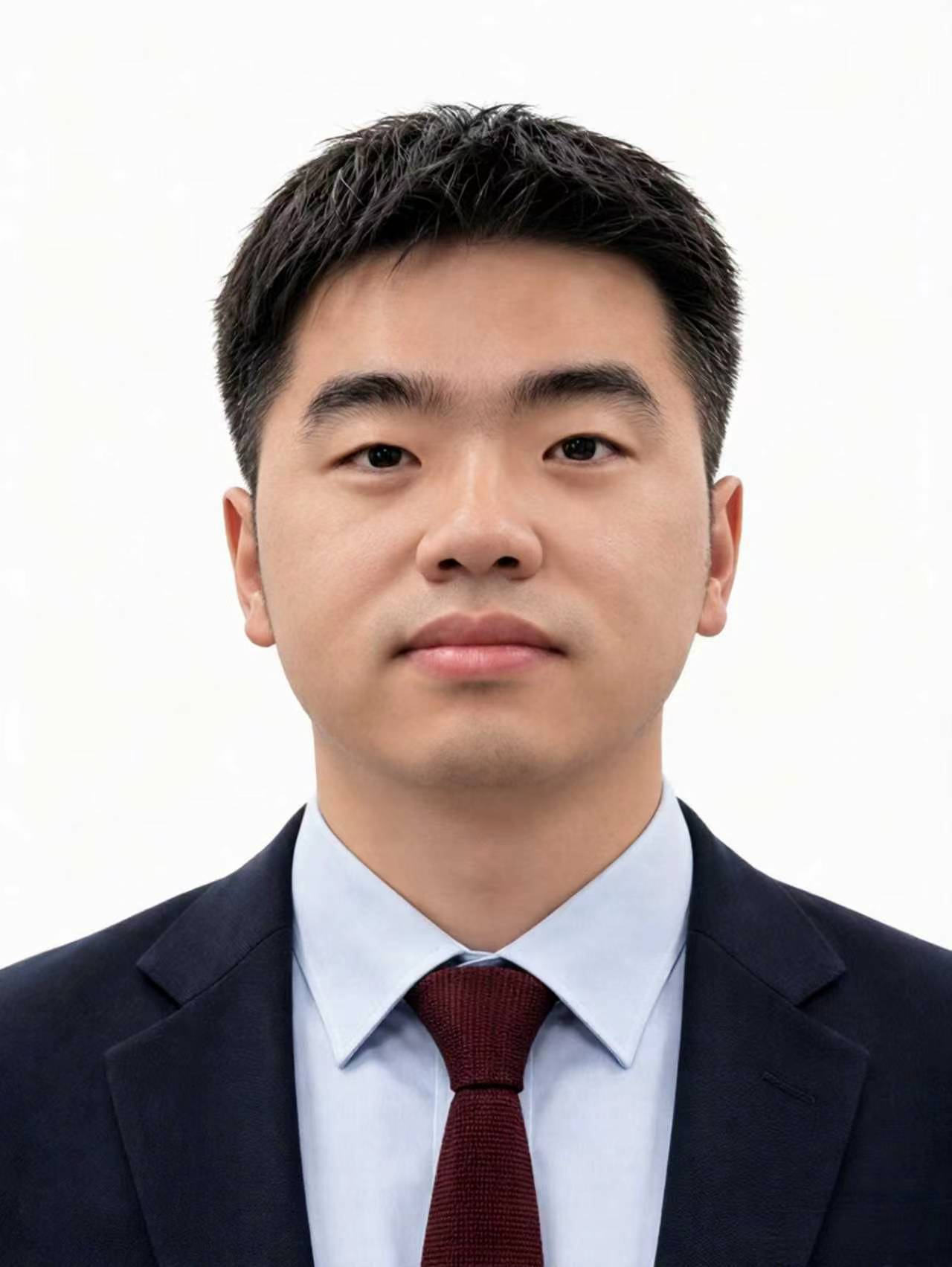}}]{Songlin~Li} is currently pursuing a Ph.D. degree in Computer Science and Technology at the School of Computer Science and Technology, Xinjiang University, China. His current research interests include co-camouflage object detection, image manipulation detection and deep learning. He has published more than 10 papers in international journals and conferences such as Proceedings of the AAAI Conference on Artificial Intelligence (AAAI), IEEE Transactions on Circuits and Systems for Video Technology (TCSVT), the IEEE International Conference on Multimedia \& Expo (ICME), and the International Conference on Multimedia Retrieval (ICMR), etc.
\end{IEEEbiography}

\begin{IEEEbiography}[{\includegraphics[width=1.0in,height=1.2in,clip,keepaspectratio]{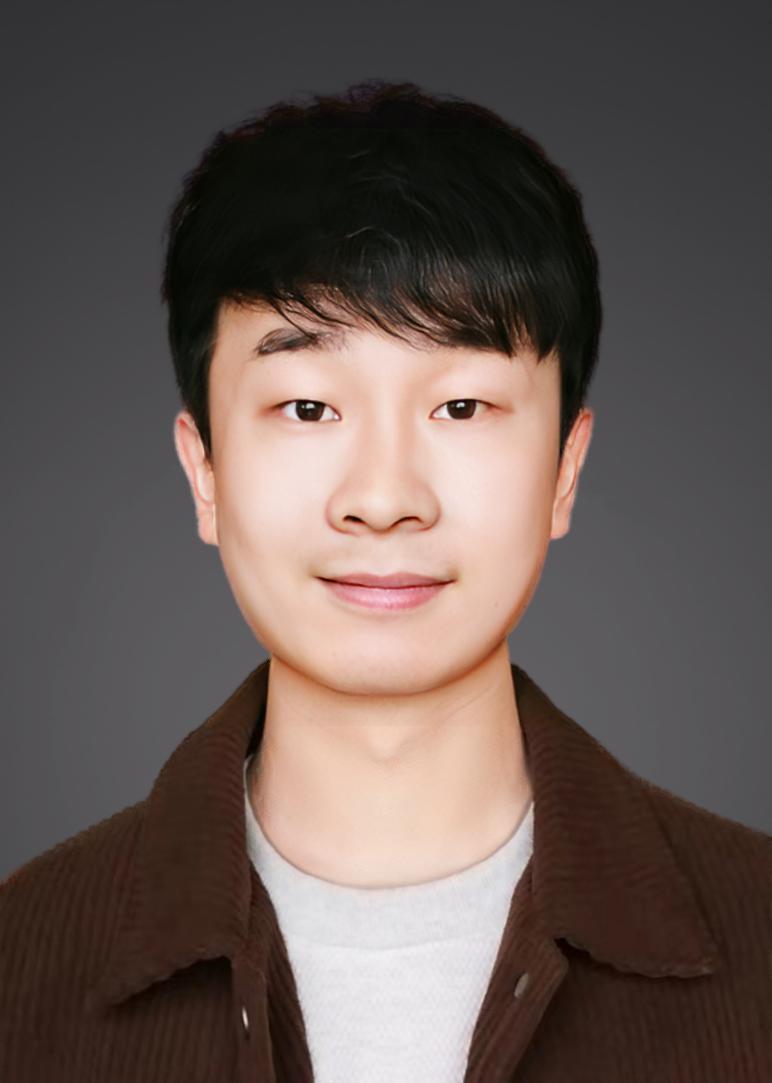}}]{Zhiqing~Guo (Member, IEEE)} received the Ph.D. degree in Computer Science and Technology from Hunan University in 2023. He is currently an Associate Professor at the School of Computer Science and Technology, Xinjiang University, Urumqi, China. At present, he is the PI of several projects, such as the Natural Science Foundation of China (NSFC), with a total fund of more than 2 million yuan. His research interests include multimedia forensics, image/video content security, visual scene understanding. He has published more than 50 papers in international journals or conferences such as IEEE Transactions on Information Forensics and Security (TIFS), IEEE Transactions on Multimedia (TMM), IEEE Transactions on Circuits and Systems for Video Technology (TCSVT), ACM Transactions on Multimedia Computing Communications and Applications (TOMM), International Conference on Machine Learning (ICML), IEEE/CVF Conference on Computer Vision and Pattern Recognition (CVPR), International Conference on Computer Vision (ICCV), Proceedings of the AAAI Conference on Artificial Intelligence (AAAI), etc.
\end{IEEEbiography}

\begin{IEEEbiography}[{\includegraphics[width=1.0in,height=1.2in,clip,keepaspectratio]{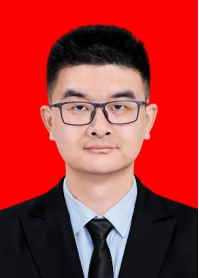}}]{Kebin~Chen} was born in 1987. He received the B.S. and M.S. degrees from Xi’an Jiaotong University, Xi’an, China, in 2011 and 2013, respectively. He received his Ph.D. degree from National University of Defense Technology, Wuhan, China in 2023. He is currently an Associate Professor at the Information Support Force Engineering University. His research interests include command information systems and visual scene understanding. He is currently the PI of a Key Project of the National Social Science Fund. He has published more than 30 papers and served as the editor‑in‑chief for one monograph.
\end{IEEEbiography}

\begin{IEEEbiography}[{\includegraphics[width=1.0in,height=1.2in,clip,keepaspectratio]{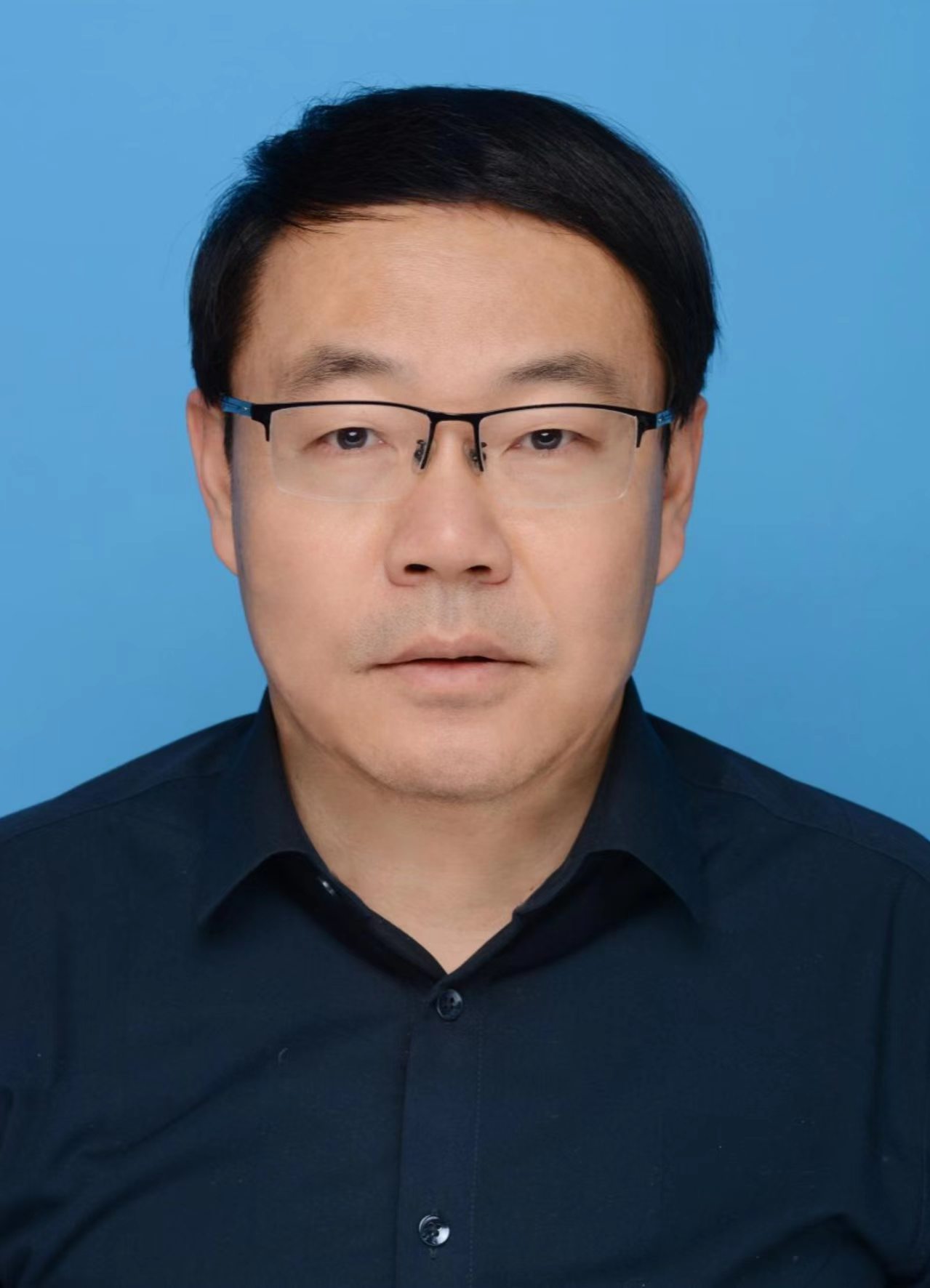}}]{Wenzhong~Yang} received his Ph.D. degree in the School of Computer Science from Wuhan University in 2011. He is currently a full professor with the School of Information Science and Engineering, Xinjiang University. He is the PI of several projects such as Natural Science Foundation of China (NSFC) and Leading Talent Project of Autonomous Region. His research interests include cyberspace security, wireless network, and machine learning.
\end{IEEEbiography}

\begin{IEEEbiography}[{\includegraphics[width=1.0in,height=1.2in,clip,keepaspectratio]{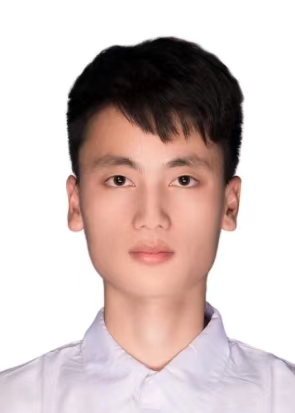}}]{Changtao~Miao} received the Ph.D. degree in Cyberspace Security from the University of Science and Technology of China in 2025. He is currently working at Ant Group, Hangzhou, China. His research interests include digital media forensics, multimodal large language model, and AI security. He has published more than 20 papers in international journals or conferences such as IEEE Transactions on Information Forensics and Security (TIFS), IEEE Transactions on Multimedia (TMM), and ACM Transactions on Multimedia Computing Communications and Applications (TOMM).
\end{IEEEbiography}

\begin{IEEEbiography}[{\includegraphics[width=1.0in,height=1.2in,clip,keepaspectratio]{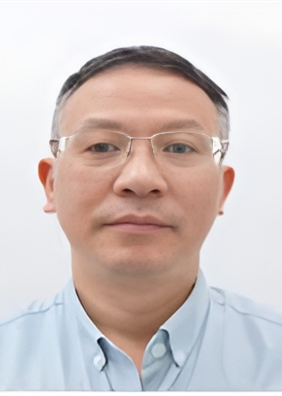}}]{Gaobo~Yang} received the Ph.D. degree in Communication and Information System from Shanghai University in 2004. He is a professor in Hunan University, China. He made an academic visit to University of Surrey, UK, from August 2010 to August 2011. He is the PI of several projects such as Natural Science Foundation of China (NSFC), Special Pro-phase Project on National Basic Research Program of China (973) and program for New Century Excellent Talents (NCET) in university. His research interests include image and video signal processing, digital media forensics. He has published research papers in international journals such as IEEE Transactions on Multimedia, IEEE Transactions on Circuits and Systems for Video Technology, IEEE Transactions on Broadcasting, ACM Transactions on Multimedia Computing, Communications, and Applications, etc.
\end{IEEEbiography}

\begin{IEEEbiography}[{\includegraphics[width=1.0in,height=1.2in,clip,keepaspectratio]{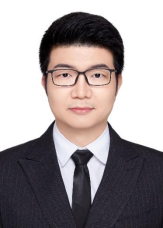}}]{Xin~Liao (Senior Member, IEEE)} received the B.E. and Ph.D. degrees in information security from the Beijing University of Posts and Telecommunications in 2007 and 2012, respectively. He is currently a Professor with the College of Cyber Science and Technology, Hunan University, where he also serves as the Vice Dean. He worked as a Post-Doctoral Fellow with the Institute of Software, Chinese Academy of Sciences, and also a Research Associate with The University of Hong Kong. From 2016 to 2017, he was a Visiting Scholar with the University of Maryland, College Park, USA. His current research interests include multimedia forensics, AI security, and watermarking. He is the Vice Chair of Technical Committee (TC) on Multimedia Security and Forensics of Asia-Pacific Signal and Information Processing Association, a member of TC on Computer Forensics of the Chinese Institute of Electronics, and a member of TC on Digital Forensics and Security of the China Society of Image and Graphics. He is serving as an Associate Editor for the IEEE Signal Processing Magazine.
\end{IEEEbiography}


\end{document}